%% file: master-document.tex
\documentclass[12pt,a4paper]{report}

\usepackage[utf8]{inputenc}
\usepackage[T1]{fontenc}

\usepackage[numbers]{natbib}

\usepackage{amsmath,amssymb,bm}

\usepackage{algorithm}
\usepackage{algorithmic}

\usepackage{booktabs}
\usepackage{array}

\usepackage{graphicx}
\usepackage[below]{placeins}
\usepackage{caption}
\usepackage{subcaption}

\usepackage{hyperref}
\usepackage{soul}
\usepackage{moreverb}
\usepackage{epstopdf}
\usepackage{psfrag}

\usepackage{amsthm}
\swapnumbers 
\theoremstyle{plain}

\theoremstyle{definition}

\numberwithin{equation}{section}

\title{Quantile-Scaled Bayesian Optimization Using Rank-Only Feedback}
\author{
  Tunde Fahd Egunjobi \\
  {\small African Institute for Mathematical Sciences, Ghana} \\
  {\small \texttt{tunde@aims.edu.gh}}
}
\date{June 2025}

\begin{document}

\maketitle

\pagenumbering{roman}

\input{abstract}

\tableofcontents
\newpage

\pagenumbering{arabic}

\input{chapter1}
\input{chapter2}

\input{chapter3}

\input{chapter4}
\input{chapter5}

\renewcommand{\bibname}{References}
\bibliographystyle{plainnat}
\bibliography{references}
\addcontentsline{toc}{chapter}{References}

\end{document}

%% file: abstract.tex
\chapter*{Abstract} 
\addcontentsline{toc}{chapter}{Abstract}

Bayesian Optimization (BO) is widely used for optimizing expensive black-box functions, particularly in hyperparameter tuning. However, standard BO assumes access to precise objective values, which may be unavailable, noisy, or unreliable in real-world settings where only relative or rank-based feedback can be obtained. In this study, we propose Quantile-Scaled Bayesian Optimization (QS-BO), a principled rank-based optimization framework. QS-BO converts ranks into heteroscedastic Gaussian targets through a quantile-scaling pipeline, enabling the use of Gaussian process surrogates and standard acquisition functions without requiring explicit metric scores. We evaluate QS-BO on synthetic benchmark functions, including one- and two-dimensional nonlinear functions and the Branin function, and compare its performance against Random Search. Results demonstrate that QS-BO consistently achieves lower objective values and exhibits greater stability across runs. Statistical tests further confirm that QS-BO significantly outperforms Random Search at the 1\% significance level. These findings establish QS-BO as a practical and effective extension of Bayesian Optimization for rank-only feedback, with promising applications in preference learning, recommendation, and human-in-the-loop optimization where absolute metric values are unavailable or unreliable.

\vspace{1cm}

\textbf{Keywords}:Bayesian Optimization; Rank-based Optimization; Quantile Scaling; Gaussian Processes; Hyperparameter Tuning; Preference Learning; Black-box Optimization. 


%% file: chapter1.tex
\chapter{Introduction}

\section{Background of the Study}

The idea of optimization has existed since the dawn of human civilization, when resources were distributed to maximize well-being \cite{garnett2023bayesian}. It cuts across many domains: in business, firms seek to maximize shareholder value; in physics, systems seek states of minimal energy; and in machine learning, models are optimized to achieve best predictive performance \cite{kochenderfer2019algorithms}. Any procedure aimed at selecting the most efficient option out of multiple options can be termed optimization.

In machine learning (ML), optimization plays a major role, especially the use of gradient descent on the loss function during model training, and tuning hyperparameters to achieve the best possible performance. Hyperparameter optimization (HPO) is the process of finding the right set of hyperparameters that enhances an ML model's performance. For example, in a neural network, this includes selecting the number of layers, type of regularization, activation functions, number of training epochs, and other model-specific configurations \cite{li2018massively,li2018hyperband}. Unlike model parameters, which are learned from data during training, hyperparameters are set before training and can significantly influence a model’s performance \cite{klein2017fast}.

Despite being an optimization problem, HPO is challenging and somewhat unusual compared to many traditional optimization problems. Most optimization problems often assume that the objective function is mathematically expressible, differentiable, convex, or at least cheap to evaluate \cite{brochu2010tutorial, frazier2018tutorial}. In contrast, the objective function in HPO, which involves training and validating a machine learning model, is often non-differentiable, non-linear, computationally expensive, and essentially a black box. The only way to find out how good any specific hyperparameter configuration is to train the model and evaluate its performance.

Early approaches, such as Grid Search, try every possible combination of hyperparameters in a predefined search space, and Random Search, which samples configurations at random \cite{bergstra2012random}, have been widely used. However, as modern machine learning models have grown in complexity and computational cost, these methods have become increasingly impractical. The increase in the number of hyperparameters, and the high cost of training model have made grid search infeasible and random search inefficient \cite{tom2024ranking}.

This has led to the adoption of \textbf{Bayesian Optimization (BO)} for hyperparameter tuning. BO is well-suited for optimizing expensive, black-box, and non-differentiable functions where evaluating the objective function $f$ is costly \cite{brochu2010tutorial, frazier2018tutorial, garnett2023bayesian}. The notion of "expensive" varies with context, it may involve the financial cost of cloud computing resources or the significant time required to train complex models, which can range from hours to days. BO addresses this by building a probabilistic surrogate model of the objective function, which it uses to intelligently select promising hyperparameter configurations to evaluate, thereby reducing the total number of expensive model training runs needed to discover good configurations \cite{frazier2018tutorial, garnett2023bayesian}.

Standard BO typically assumes continuous, scalar-valued objective functions, where each configuration returns a numeric performance metric (such as accuracy or loss) upon evaluation. However, in many practical scenarios, especially in situations where only qualitative assessments or ranking information are available, obtaining a precise metric is difficult or impossible \cite{nguyen2021top}. Also, in some applications where the absolute metric value is noisy, unreliable, or not as important as the relative ordering of configurations, it may be preferable to rely on ordinal comparisons rather than exact performance scores \cite{tom2024ranking}. For example, in hyperparameter tuning for recommendation systems or ranking models, it might be sufficient to know which configuration yields better user engagement or ranking quality without requiring precise numerical metrics.

These limitations motivate the study of Rank-Based Bayesian Optimization (RBO), which relies on ordinal feedback rather than explicit metrics. RBO preserves the sample efficiency of BO but relaxes the feedback assumption, making it applicable to problems where only relative performance is observable. The present study develops and evaluates such a method, demonstrating its feasibility in synthetic settings relevant to hyperparameter tuning.

\section{Problem Statement}

In many real-world optimization problems, the absolute values of objective functions are noisy, unreliable, or unavailable; for example, it will be difficult to assign a numerical value to the taste of two foods, but easier to know which tastes better. Also, in a user preference study, ranking will prevail over numerical scores \cite{nguyen2021top, tom2024ranking}. Ranking provides a more intuitive understanding of preferences, allowing researchers and decision-makers to focus on relative comparisons rather than exact measurements. This approach will enhance the practicality of BO and lead to more effective algorithms that capture the nuances of human judgment, ultimately enhancing the outcomes of optimization tasks.

Several methods have been developed and used for BO over the years, with only very few studies and methods available for RBO. To bridge this gap, in this study, we propose a rank-based BO method that only requires the rank and not the absolute score of the function values. 

That is, for any given function $f$, only the relative order of $f$

\[ f(x_1)<f(x_2)< ... < f(x_n),\]

will be required rather than the exact value of $f$. 

\section{Research Aim and Objectives}

The aim of this study is to develop and evaluate a Quantile-Scaled Bayesian Optimization (QS-BO) method for black-box optimization problems when only rank information is available.

The objectives of the study are to:

\begin{itemize}
    \item Propose a quantile-scaling framework that converts rank information into heteroscedastic Gaussian targets suitable for Gaussian process modeling.
    \item Implement an iterative Bayesian optimization framework that leverages rank-only feedback.
    \item Benchmark QS-BO against a conventional non-BO baseline (Random Search) on synthetic benchmark functions.
\end{itemize}

\section{Scope and Limitation of the Study}

This study is limited to the development and empirical evaluation of Quantile-Scaled Bayesian Optimization (QS-BO) in settings where only rank information is available. The experiments are conducted exclusively on synthetic benchmark functions that are widely used in the Bayesian optimization literature, namely a one-dimensional nonlinear function, the Forrester function, and the two-dimensional Branin function. These functions are selected because they exhibit non-trivial landscapes that require balancing exploration and exploitation.

The study does not extend to real-world machine learning tasks or large-scale optimization problems. While QS-BO is motivated by potential applications in hyperparameter tuning and other practical domains where only rank information may be available, such applications fall outside the scope of this work. Furthermore, comparisons are limited to Random Search as a non-BO baseline. Other rank-based methods, as well as standard Bayesian optimization frameworks operating on metric scores, are acknowledged but not comprehensively evaluated here.

The primary focus is on demonstrating the feasibility, statistical foundation, and optimization dynamics of QS-BO in controlled synthetic settings. Future work may extend the method to real-world problems, larger domains, and comparisons against a wider range of baselines.

\section{Structure of the Thesis}

The remainder of the thesis is structured as follows:

\begin{itemize}
    \item Chapter two focuses on the theoretical background of Bayesian optimization and reviews past studies on rank-based Bayesian optimization.
    \item Chapter three describes our proposed QS-BO method, including the statistical modeling choices, acquisition functions, and implementation details.
    \item Chapter four presents experiments benchmarking QS-BO against Random Search on synthetic functions, along with statistical comparisons and visualizations of optimization dynamics.
    \item Chapter five concludes the study and suggests directions for future research.
\end{itemize}

%% file: chapter2.tex
\chapter{ Theoretical Foundations and Related Work}

This chapter reviews the theoretical foundation of standard BO and highlights existing work in rank-based models. It begins with an overview of the BO and its standard workflow, explaining the commonly used surrogate model (Gaussian Process) and acquisition function that make up the BO. We then proceed to the Tree-Structured Parzen estimator approach, which is often considered in high-dimensional BO. Finally, we review papers that have adopted similar or related techniques in rank-based Bayesian Optimization and provide an overview of the methods explored in this area.

\section{Overview of Bayesian Optimization}

Since the 1960s, the BO approach has been progressively studied by statistics and machine learning communities \cite{garnett2023bayesian}. It gained popularity through the study of Jones et al., (1998) \cite{jones1998efficient} on Efficient Global Optimization and has been widely used across different fields where optimization is required, specifically where the objective is not very expressive or costly to evaluate, commonly referred to as “black-box” functions \cite{garnett2023bayesian, frazier2018tutorial,jones1998efficient}  


The term Bayesian was derived from the Bayes theorem in statistics because the optimization approach combines a prior distribution $p(f)$ with observed data $(\mathcal{D})$ to obtain a posterior distribution $p(f \mid \mathcal{D})$ over the space of unknown functions $f$; using the Bayes theorem \cite{brochu2010tutorial}:

\begin{align}
p(f \mid \mathcal{D}) \propto p(\mathcal{D} \mid f) \, p(f).
\end{align}

\begin{figure}[htbp]
    \centering
    \includegraphics[width=0.7\textwidth]{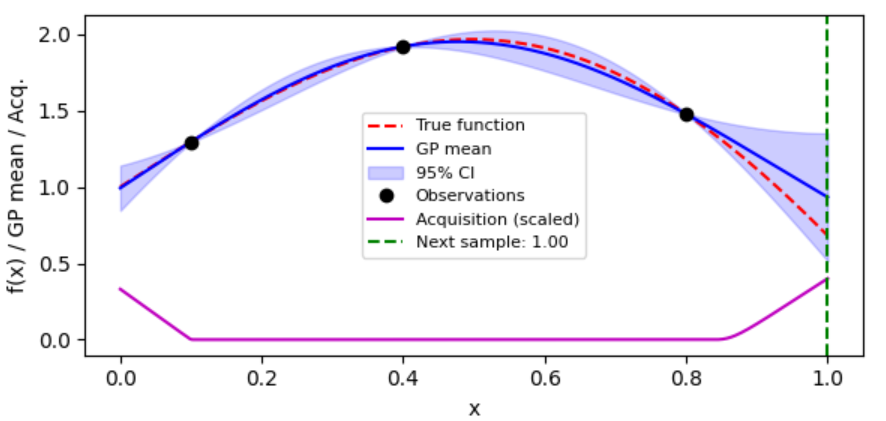}
    \caption{Bayesian Optimization process showing the Gaussian Process posterior mean and variance, the acquisition function below, and selected observation points. The arrow (green line) indicates the next sampling location suggested by the acquisition function.}
    \label{fig:bo_process}
\end{figure}

The general BO framework, be it one or multi-objectives, multi-fidelity, preference-based or rank-based, etc, has always been to select a prior distribution that will model the true objective function, often referred to as a surrogate model, and the acquisition function that will direct the posterior distribution to the next promising data point to be considered toward the goal of the optimization \citep{brochu2010tutorial,frazier2018tutorial, snoek2012practical}.  Fig. \ref{fig:bo_process} uses a Gaussian Process as a surrogate model with the blue line representing the mean function and the 95\% CI accounting for the variance. The red line represents the main function that we want to minimize, which the GP is emulating. The green line represents the acquisition function, which points to the next point to be considered. It can be observed that at point $x = 1$, the surrogate and the main function have high variation.   

\vspace{0.6cm}
\subsection{Gaussian Process}

Gaussian Process (GP) regression is a probabilistic technique for modeling function \citep{frazier2018tutorial} and has been widely used as a surrogate model for standard BO, where the set of observations (function evaluations) is continuous \citep{brochu2010tutorial, frazier2018tutorial, rasmusen2006gaussian}. A GP prior will be defined over $f$ 
\begin{align}
f(x) &\sim \mathcal{GP}(m(x), k(x, x')),
\end{align}

where 
\begin{itemize}
\item $m(x)$: is the mean function (often assumed to be zero)
\item $k(x, x')$: is the covariance (Kernel) function between any two data points. 
\end{itemize}

Given a set of observations \( \mathcal{D} = \{(X, \mathbf{y})\} \), where \( X = [x_1, \dots, x_n] \) and \( \mathbf{y} = [y_1, \dots, y_n]^T \), and assuming a Gaussian likelihood, to obtain the posterior distribution, we will constraint the prior distribution to include only those functions that are consistent with the observed data points. i.e., we update the prior with new input points \( X_* \).

\begin{align}
f(X_*) \mid X, \mathbf{y}, X_* &\sim \mathcal{N}(\mathbf{\mu}_*, \mathbf{\Sigma}_*)
\end{align}

where
\begin{align}
\mathbf{\mu}_* &= K(X_*, X) \left[K(X, X) + \sigma^2_n I\right]^{-1} \mathbf{y}
\end{align}
and
\begin{align}
\mathbf{\Sigma}_* &= K(X_*, X_*) - K(X_*, X) \left[K(X, X) + \sigma^2_n I\right]^{-1} K(X, X_*).
\end{align}

Here, \( K(A, B) \) denotes the covariance matrix computed using the kernel function \( k(\cdot, \cdot) \) between inputs \( A \) and \( B \), \( \sigma^2_n \) is the noise variance, and \( I \) is the identity matrix \citep{rasmusen2006gaussian}. The additional part $\sigma^2_nI$ is added to the covariance function to account for the noisy observation that is often observed in reality. In a realistic situation, we often don't observe the actual function value $f$ but a noisy value $y = f + e$ where $e \sim \mathcal{N}(0, \sigma^2)$ accounts for the noise. The choice of the kernel function $k(.,.)$ and its hyperparameters is another factor in GP. The most preferred kernel functions are the Square-Exponential and Matáern kernel \cite{frazier2018tutorial, wang2023recent}. For full details on Gaussian process, read Rasmussen and Williams (2006) \citep{rasmusen2006gaussian}.

\vspace{0.6cm}
\subsection{Acquisition Functions}

Since the goal of optimization is to avoid many evaluations of the true function because of the cost or its nature. The acquisition function (AF) provides guidance on the next promising points to be evaluated.
Given the current best value $x_{best}$, the acquisition function dictates the next input $x_{next}$ to be considered, which is expected to yield a better result than the current $x_{best}$.  There are several choices of AF: Probability of Improvement, Expected Improvement, Entropy Search, Knowledge Gradient, and so on  \citep{brochu2010tutorial,frazier2018tutorial}.

Let us assume a minimization problem, then $f_{best} = \text{min} f$, that is, the minimum observation so far. 
\begin{itemize}
\item \textbf{Probability of Improvement (PI)}: It quantifies how likely it is for a new point to perform better than the $x_{best}$ \cite{frazier2018tutorial}

\begin{align}
\textbf{PI}(x) = \mathbb{P}(f(x)\le f_{best}) = \Phi\left(\frac{f_{best} - \mu(x) }{\sigma(x)}\right).
\end{align}
PI is known for exploitation; it underexplores the whole region, which sometimes makes it get stuck in the local optimum. 

\item \textbf{Expected Improvement (EI)}: This is the most commonly used AF \cite{zhan2020expected, wang2023recent}, while PI focuses on how likely a point is to improve the current best, EI measures the amount of increment expected if a point is to be considered. It chooses the next data point that has the highest expected improvement \cite{zhan2020expected,frazier2018tutorial}. It is given as:

\begin{equation}
\text{EI}(x) = \mathbb{E}\left[ \max(f_{best} - f(x), 0) \right].
\end{equation}
It has a closed-form formula for a surrogate GP in relation with the standard normal distribution as:
\begin{equation}
{EI}(x)= (f_{best} - \mu(x)) \, \Phi\left( \frac{f_{best} - \mu(x)}{\sigma(x)} \right) 
+ \sigma(x) \, \phi\left( \frac{f_{best} - \mu(x)}{\sigma(x)} \right),
\end{equation}

where  $\Phi (Z)$ and $\phi(Z)$ are the CDF anf PDF of a standard normal distribution and $\mu$ and $\sigma$ are the GP predictive mean and standard deviation at point $x$.

\item \textbf{Knowledge Gradient (KG)}: PI and EI focus on how much better a point can be than the current one, but KG focuses on how much I will benefit in the long run by evaluating this point. It rewards the points that help us learn about the main function globally, rather than just immediate improvement. KG selects the next evaluation point by estimating the expected gain in the best-so-far value after sampling at a new point. It follows the value of information principle, evaluating how learning at \( x \) improves future decisions. Given a Gaussian Process, the predictive distribution is normal, allowing KG to be computed as:

\begin{align}
KG(x) = \mathbb{E}\left[\max\left(f(x), f_{best}\right)\right] - f_{best}.
\end{align}
Unlike Expected Improvement, KG explicitly accounts for the information gained by sampling at \( x \) \cite{frazier2018tutorial}.
\end{itemize}

\subsection{Tree-structured Parzen Estimator (TPE) Approach}

Generally, in BO we model $p(y \mid x)$ i.e. performance given parameters especially when working with GP but TPE do the direct opposite, it models $p(x|y)$ i.e., parameter given performance \cite{bergstra2011algorithms}. It models $p(x \mid y)$ using two probability density functions:
\begin{align}
p(x \mid y) = 
\begin{cases}
\ell(x), & \text{if } y < y^* \\
g(x), & \text{if } y \geq y^*,
\end{cases}
\end{align}

where $\ell(x)$ and $g(x)$ are considered as good and bad samples, respectively. TPE uses density estimator (like kernel density estimator) to estimate the $\ell(x)$ and $g(x)$ \cite{ozaki2022multiobjective, bergstra2011algorithms}. $\ell(x)$ is derived from the set of observations $(X)$ whose performance are better (below when minimizing) than $y^*$ i.e. $p(x \mid y < y^*)$ and $g(x)$ is derived from the rest of the observations. Both $\ell(x)$ and $g(x)$ can be considered as surrogate models for the good and bad observations, given their performance. $y^*$ serves as a performance threshold which is chosen based on $\gamma$-quantile of the $y$ values, where $\gamma$ controls the fraction of the best-performing trials to be considered good \cite{watanabe2023tree}.

To select the next promising observation $x_{next}$, TPE adopts EI acquisition function which has been shown by \cite{bergstra2011algorithms} that it is proportional to the ratio of good to bad:

\begin{align}
\mathbb{EI}(x) \propto \frac{\ell(x)}{g(x)}.
\label{TPE_EI}
\end{align} 

\begin{figure}[htbp]
    \centering
    \includegraphics[width=1\textwidth]{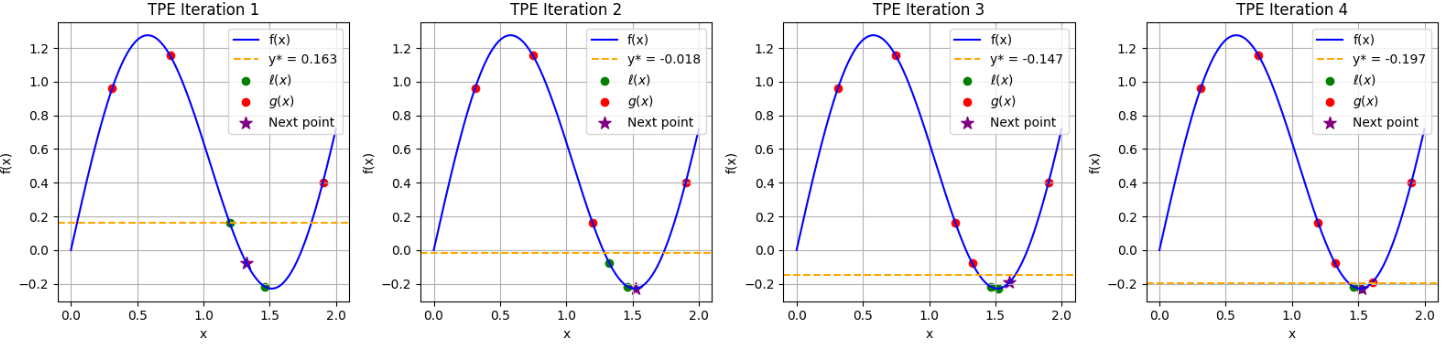}
    \caption{TPE Approach showing how the $y^*, \ell(x), g(x)$ as well as the next point selected by $\ell(x)/g(x)$ changes with iteration. The iterations goes on till the allocated budget is exhausted.}
    \label{fig:TPE}
\end{figure}

The expression \eqref{TPE_EI} implies that for a data point $x_{next}$ to be selected, it must have high probability under $\ell(x)$ and low under $g(x)$. Figure \ref{fig:TPE} demonstrates TPE on a one-dimensional function, demonstrating the transitioning of data points from good to bad samples as the iteration progresses.

\section{Past Studies on Rank-Based Bayesian Optimization}
Several studies have been conducted on standard BO, sometimes referred to as conventional BO, but only a few studies have been conducted on Rank-Based BO (RBO). Although Preferential Bayesian Optimization (PBO) has also been explored severally by many researchers. PBO also uses rank rather than exact function values. In PBO, the objective function $f$ is a latent utility function defined on individual input $x$, but since the exact function values are not observed, the algorithm is designed to query pairs of inputs $(x_i, x_j)$ and observe the preference between them, returning an output in the form of $f(x_i)>f(x_j)$ when $x_i$ is preferred to $x_j$ or otherwise. 
\vspace{0.5cm}

\subsection{Preferential Bayesian Optimization (PBO)}


\textbf{Problem Setup:}
The setup is designed like every other optimization problem. There will be an objective function $f$ which we wish to minimize or maximize. Suppose that we want to maximize $f$, that is, we are searching for an optimal point $x^*$, such that  \[
x^\star = \arg\max_{x \in \mathcal{X}} f(x).
\]

We define:
\begin{itemize}
    \item \(\mathcal{X} \subset \mathbb{R}^d\) be the input space.
    \item $x$ is a data point in $X$ $(x \in X)$.
    \item \(f: \mathcal{X} \rightarrow \mathbb{R}\) be an unknown latent utility function we wish to maximize.
\end{itemize}

But instead of observing the exact function values, we only obtain preferences between pairs of points.

Specifically, given a pair \((x_i, x_j)\), we observe
\[
y_{ij} =
\begin{cases}
1 & \text{if } f(x_i) + \epsilon_i > f(x_j) + \epsilon_j \\
0 & \text{otherwise}
\end{cases}
\]
where \(\epsilon_i, \epsilon_j \sim \mathcal{N}(0, \sigma^2)\) are independent Gaussian noise terms.

PBO workflows follow the same pattern as the standard BO, but only incorporate the preference that is observed rather than the exact values. The changes are corrected in the likelihood function. Rather than a complete Gaussian likelihood, PBO uses the Thurstone (Probit) model or the Bradley-Terry model to accommodate the pairwise comparison between two points $(x_i, x_j)$, which are then incorporated into the posterior \cite{chu2005preference}. The AF can be Expected Improvement for preference, Thompson sampling, etc.  \\

\textbf{1. GP Prior:}
\[
f \sim \mathcal{GP}(m(x), k(x, x')).
\]

\textbf{2. Preference Likelihood:}

Given a pairwise preference observation \(y_{ij}\) where
\[
y_{ij} = 
\begin{cases}
1 & \text{if } f(x_i) + \epsilon_i > f(x_j) + \epsilon_j \\
0 & \text{otherwise}
\end{cases}
\]
with \(\epsilon_i, \epsilon_j \sim \mathcal{N}(0, \sigma^2)\),  
the likelihood can be modeled by either:

- \textbf{Thurstone (Probit) model:}
\[
P(y_{ij} = 1 \mid f) = \Phi\left(\frac{f(x_i) - f(x_j)}{\sqrt{2}\sigma}\right)
\]

or

- \textbf{Bradley-Terry (Logit) model:}
\[
P(y_{ij} = 1 \mid f) = \frac{1}{1 + \exp\left(-\frac{f(x_i) - f(x_j)}{\sigma}\right)}
\]

\textbf{3. Posterior Inference:}
\[
p(f \mid \mathcal{D}_t) \propto p(f) \prod_{(i,j)} P(y_{ij} \mid f)
\]

\textbf{4. Acquisition Function:}
\[
(x_{t+1}, x'_{t+1}) = \arg\max_{(x, x')} a(x, x'; \mathcal{D}_t)
\]

\textbf{5. Recommendation (Current Best Guess):}
\[
x^\star_t = \arg\max_{x \in \mathcal{X}} \mathbb{E}[f(x) \mid \mathcal{D}_t].
\]

While the PBO algorithm has been designed to query pairs of input, Nguyen et al., \cite{nguyen2021top} generalized PBO to any $k\in \mathbb{R}$ sets of inputs in their Top-k Ranking BO paper. In their work, they introduced a novel technique to accommodate top-k ranking $(x_1, ..., x_k)$, which serves as an extension and a generalized PBO with pair input $(x_i, x_j)$. The extension from pair input to $k$ length differentiates PBO from RBO, which makes their work directly related to this study.  They designed a surrogate model that is inspired by the Multinomial Logit Model, which is capable of handling top-k ranking and ties. They define their preference as  

\begin{align}
  p\left(x > X' \setminus \{x\}, f_{X' \cup \{x\}}; \delta \right)
= \frac{\exp(f(x))}{\exp(f(x)) + \sum_{x' \in X' \setminus \{x\}} \exp(f(x')) + \delta}  
,
\end{align}

which represents the probability that $x$ is preferred over a set of inputs $X' \subset X$. The value $\delta$ handles the ties that may be observed if two values are not better than each other. When $\delta=0$, it is considered the generalized BT model, which is also known as the Plackett-Luce model. They also introduced  Multinomial Predictive Entropy Search (MPES), the first information-theoretical AF designed for PBO. MPES is based on information theory for its selection of the next query point. To conclude their study, they demonstrated the superiority of their proposed methods (MPES) over other existing AF, like EI and Discrete Thompson Sampling, with the CIFAR10 dataset, which is often used for benchmarking in the field of BO.

The other work on RBO is the work of Tom et al., \cite{tom2024ranking} on Ranking over Regression for Bayesian Optimization and Molecule Selections. They highlighted that the relative order of the molecules is more important than the exact value that will be obtained when regression-based BO (standard BO) is used. Hence, they adopted ranking models rather than the regression model for their surrogate models. Although the work did not develop any new technique, they replaced the standard regression-based surrogate model (GP) with ranking models like Multilayer Perceptron (MLP), Graph Neural Network (GNN), and Bayesian Neural Network (BNN) and trained them with ranking loss in place of regression loss. They demonstrated that ranking over regression provided the same or an improved performance on their rough chemical dataset, where the relative ordering is more important than exact values. The main information that this paper contributes to our study is the belief that in a noisy or rough dataset. RBO performance will either match or outperform the regression-based BO, supporting the necessity to study RBO.


%% file: chapter3.tex
\chapter{Proposed Method}

We propose a simple, principled strategy for Bayesian optimization when only \emph{rank} (ordinal) feedback is available. The method maps ranks to Gaussian pseudo-observations, injects a principled heteroscedastic noise model derived from order-statistic theory, and then uses a standard Gaussian Process (GP) surrogate with off-the-shelf acquisitions (EI/UCB/TS). The advantages are that the method (i) is invariant to monotone transformations of the objective, (ii) is calibration-free, and (iii) plugs directly into existing BO pipelines.

\section{Mathematical Background and Rationale}

This section collects the probabilistic facts used by the method and gives a short rationale for each.

\subsection{Uniform distribution and normalized ranks}
\leavevmode

Let \( \{x_i\}_{i=1}^n \subset \mathcal{X}\) be the points evaluated so far and let \(r_i\in\{1,\dots,n\}\) denote the rank of point \(x_i\) among those \(n\) evaluations (we use \(r=1\) for the best point in the minimization setting). We convert a rank \(r\) to a point on the unit interval by
\begin{equation}
u \;=\; \frac{r - 0.5}{n}.
\label{eq:u_def}
\end{equation}
 Normalizing ranks to \([0,1]\) gives a natural link to order-statistics and to quantile transforms. The continuity correction \( -0.5 \) (or, alternatively, \(r/(n+1)\)) prevents mapping ranks to the exact boundaries \(0\) or \(1\) (which would produce \(\Phi^{-1}(0)\) or \(\Phi^{-1}(1)\)). The result \(u\) is interpreted as an empirical quantile associated with that rank.

\subsection{Order statistics and the Beta distribution}

If \(U_1,\dots,U_n\) are i.i.d.\ \(\mathrm{Uniform}(0,1)\) and \(U_{(1)}\le \dots \le U_{(n)}\) are the order statistics, then the \(k\)-th order statistic has a Beta distribution \cite{david2004order}:
\begin{equation}
U_{(k)} \;\sim\; \mathrm{Beta}(k,\; n-k+1).
\label{eq:order_beta}
\end{equation}
In particular,
\[
\mathbb{E}[U_{(k)}] = \frac{k}{n+1},\qquad
\mathrm{Var}[U_{(k)}] = \frac{k(n-k+1)}{(n+1)^2(n+2)}.
\]
When all we know about a point is its rank among \(n\), its true underlying quantile on \([0,1]\) is not a point mass but has distribution given by the Beta law above. Treating the rank as the \(k\)-th order statistic under a uniform reference model gives a principled, closed form for the uncertainty of the quantile estimate \cite{david2004order, gentle2009computational}.

\subsection{Gaussian quantile (probit) transform}

To obtain targets suitable for GP regression (which naturally handles Gaussian noise), we transform the uniform quantile \(u\) \eqref{eq:u_def} to the Gaussian scale via the inverse standard normal CDF:
\begin{equation}
z \;=\; \Phi^{-1}(u).
\label{eq:z_def}
\end{equation}
Here \(\Phi^{-1}\) is the standard normal quantile function and \(\phi\) denotes the standard normal density.

Mapping to the Gaussian scale yields targets that are approximately Gaussian when the underlying quantile is near its mean (by the Central Limit intuition) and — crucially — allows us to use standard GP regression machinery without custom preference likelihoods. The Gaussian scale is also convenient because the Delta method (below) has a simple form.

\subsection{Propagating Beta uncertainty to the z–scale: delta method}

The Beta distribution for \(U_{(k)}\) supplies a variance on the \(u\)-scale. To obtain an approximate variance for \(Z=\Phi^{-1}(U)\) we apply the delta method. If \(g(u)=\Phi^{-1}(u)\) then
\[
\mathrm{Var}\big(g(U)\big) \approx \big(g'( \mathbb{E}[U]))\big)^2 \mathrm{Var}(U).
\]
Since \(g'(u) = 1/\phi(g(u))\), for the observed \(u_i\) and \(z_i=\Phi^{-1}(u_i)\) we obtain:
\begin{equation}
\sigma_{z,i}^2 \;\approx\; \frac{\mathrm{Var}(U_{(r_i)})}{\phi(z_i)^2}
\;=\;
\frac{r_i (n+1-r_i)}{(n+1)^2(n+2)} \cdot \frac{1}{\phi(z_i)^2}.
\label{eq:varz}
\end{equation}

This produces a heteroscedastic variance on the \(z\)-scale that is largest near the extremes (where \(\phi(z)\) is small) and reflects the combinatorial uncertainty of the order statistic. Passing these variances into the GP prevents overconfidence in points whose rank is intrinsically less informative and used to achieve robustness \cite{le2005heteroscedastic, griffiths2021achieving}.


\subsection{Delta vs.\ treating ranks as deterministic (delta / point mass)}

An alternative is to treat the rank as giving a point estimate \(u_i\) (a Dirac / delta assumption). That is equivalent to setting \(\mathrm{Var}(U_i)=0\), leading to \(\sigma_{z,i}^2=0\). This simplifies implementation but ignores the natural uncertainty induced by finite \(n\). The delta (point-mass) approach is simpler, but it ignores heteroscedastic uncertainty: for example, being second best among \(n=5\) is much noisier than being second best among \(n=100\). Equation \eqref{eq:varz} captures that dependence.

\section{Rank–based GP surrogate and acquisition}

\subsection{GP likelihood with per-point noise}
\leavevmode

Collect the transformed targets \(\mathbf{z}=[z_1,\dots,z_n]^\top\) and the corresponding noise variances \(\sigma_{z,1}^2,\dots,\sigma_{z,n}^2\). Place a Gaussian process prior on a latent function \(g:\mathcal{X}\to\mathbb{R}\),
\[
g(\cdot)\sim\mathcal{GP}(0,k_\theta(\cdot,\cdot)),
\]
and model observations as
\[
z_i = g(x_i) + \varepsilon_i,\qquad \varepsilon_i\sim\mathcal{N}(0,\sigma_{z,i}^2).
\]
Let \(K\in\mathbb{R}^{n\times n}\) be the kernel matrix with entries \(K_{ij}=k_\theta(x_i,x_j)\) and \(\Sigma=\mathrm{diag}(\sigma_{z,1}^2,\dots,\sigma_{z,n}^2)\). The marginal log-likelihood is
\begin{equation}
\log p(\mathbf{z}\mid X,\theta)
= -\tfrac12 \mathbf{z}^\top (K+\Sigma)^{-1}\mathbf{z}
  -\tfrac12\log\det(K+\Sigma) - \tfrac{n}{2}\log(2\pi).
\label{eq:gp_marginal}
\end{equation}

This is standard GP regression but with \(\Sigma\) replacing the usual scalar noise variance. The heteroscedastic diagonal \(\Sigma\) implements the rank-dependent per-point uncertainty we derived in \eqref{eq:varz}.

\subsection{GP predictive distribution}
\leavevmode

For a test point \(x_\star\) let \(k_\star=[k_\theta(x_\star,x_1),\dots,k_\theta(x_\star,x_n)]^\top\). The posterior mean and variance on the \(z\)-scale are
\begin{align}
\mu_\star &= k_\star^\top (K+\Sigma)^{-1}\mathbf{z}, \label{eq:gp_mean}\\
s_\star^2 &= k_\theta(x_\star,x_\star) - k_\star^\top (K+\Sigma)^{-1} k_\star. \label{eq:gp_var}
\end{align}
The GP gives us a predictive Gaussian belief about the latent rank-score at any candidate \(x\). Note that both mean and variance depend on \(\Sigma\): points with large \(\sigma_{z,i}^2\) contribute less to the posterior (they are downweighted).

\subsection{Acquisition functions on the latent scale}
We compute acquisition functions on the \(z\)-scale using \(\mu(x)\) and \(s(x)\). For example, Expected Improvement (minimization) is
\begin{equation}
\mathrm{EI}(x) \;=\; s(x)\Big(\phi(\gamma(x)) + \gamma(x)\Phi(\gamma(x))\Big),\qquad
\gamma(x)=\frac{z_\star - \mu(x)}{s(x)},
\label{eq:ei}
\end{equation}
where \(z_\star=\min_i z_i\) (or \(\max\) for maximization).

\paragraph{Rationale.} EI balances \emph{exploitation} (points with small \(\mu\)) and \emph{exploration} (points with large \(s\)) \cite{frazier2018tutorial, snoek2012practical}. Because the mapping rank \(\mapsto z\) is monotone, maximizing EI on the \(z\)-scale targets points that are likely to improve the rank of the current best observed configuration in terms of the original objective.

\subsection{Monotonicity and correctness of the optimizer}
If \(h:\mathbb{R}\to\mathbb{R}\) is strictly monotone, then
\[
\arg\min_x f(x) = \arg\min_x h(f(x)).
\]
The maps we use (rank \(\to\) quantile \(\to\) \(\Phi^{-1}\)) are monotone in the ordering induced by \(f\), so the optimizer of the latent GP mean is consistent (in ordering) with the optimizer of the true objective.

This is the formal reason rank-based optimization can find the true optimizer: ranks preserve the order of \(f\), and any strictly monotone transform preserves the optimizer.

\section{QS-BO algorithm (complete method)}

\begin{algorithm}[t]
\caption{QS-BO: Quantile-Scaled Bayesian Optimization}
\label{alg:qsbo}
\begin{algorithmic}[1]
\STATE \textbf{Input:} objective \(f\), search domain \(\mathcal{X}\), initial budget \(n_0\), BO iterations \(T\).
\STATE \textbf{Initialize:} sample \(n_0\) points \(x_1,\dots,x_{n_0}\) from \(\mathcal{X}\); evaluate \(y_i=f(x_i)\); compute ranks \(r_i\).
\FOR{t = \text{$n_0+1$} to \text{$n_0+T$}}

  \STATE \(\,u_i \leftarrow (r_i-0.5)/n\) \quad (map ranks to \([0,1]\)).\\
    \hfill\(\triangleright\) \emph{Rationale: prevents mapping to exact boundaries.}
  \STATE \(\,z_i \leftarrow \Phi^{-1}(u_i)\) \quad (Gaussian pseudo-observations).\\
    \hfill\(\triangleright\) \emph{Rationale: use GP-friendly targets.}
  \STATE Compute \(\mathrm{Var}(U_{(r_i)})\) via Beta formula and \(\sigma_{z,i}^2\) using \eqref{eq:varz}.\\
    \hfill\(\triangleright\) \emph{Rationale: quantify rank uncertainty; produce heteroscedastic noise.}
  \STATE Fit GP on \((x_i,z_i)\) with diagonal noise \(\Sigma=\mathrm{diag}(\sigma_{z,i}^2)\).\\
    \hfill\(\triangleright\) \emph{Rationale: downweight uncertain ranks in posterior.}
  \STATE Build candidate set \(X_{\mathrm{cand}}\) (grid or random samples), compute \(\mu, s\) from GP and acquisition \(\alpha(x)\) (e.g.\ EI).\\
    \hfill\(\triangleright\) \emph{Rationale: cheap approximate maximization of the acquisition.}
  \STATE Select \(x_t \leftarrow \arg\max_{x\in X_{\mathrm{cand}}} \alpha(x)\) and evaluate \(y_t=f(x_t)\).\\
    \hfill\(\triangleright\) \emph{Rationale: pick the candidate that best trades off exploration/exploitation.}
  \STATE Update dataset: append \(x_t, y_t\); recompute ranks for all points.
\ENDFOR
\STATE \textbf{Output:} best observed point \(\hat{x} = \arg\min_i y_i\) (or \(\arg\max\) if maximizing).
\end{algorithmic}
\end{algorithm}

\section{Practical implementation details and numerical considerations}

\begin{itemize}
  \item \textbf{Clipping/guards.} Clip \(u\) to \([\epsilon,1-\epsilon]\) (e.g.\ \(\epsilon=10^{-6}\)) to avoid \(\Phi^{-1}(0)\) or \(\Phi^{-1}(1)\). Also floor \(\phi(z)\) when computing \eqref{eq:varz} to avoid division by extremely small numbers.
  \item \textbf{Choice of kernel.} RBF or Mat\'ern kernels are suitable. Use ARD lengthscales in multivariate problems.
  \item \textbf{GP library support.} In scikit-learn, pass the heteroscedastic variances as `alpha=` (an array of length \(n\)). This performs exact GP regression with diagonal noise \(\Sigma\).
  \item \textbf{Candidate search for acquisition.} In 1D/2D you may use grids; in higher dimensions, use random or quasi-Monte Carlo candidate sets.
\end{itemize}



\section{Summary}
We presented Quantile-Scaled Bayesian Optimization (QS-BO): a drop-in rank-to-Gaussian pipeline that converts ranks into heteroscedastic Gaussian targets, fits a GP surrogate, and uses standard acquisitions for BO. Each modeling choice is justified: the Beta/order-statistic model quantifies rank uncertainty, the delta method yields a principled heteroscedastic noise model, and GP regression exploits that structure to trade off exploration and exploitation in a familiar framework.


%% file: chapter4.tex
 \chapter{Experiment and Discussion}


In our empirical evaluation, we benchmarked the performance of our proposed 
Quantile-Scaled Bayesian Optimization (QS-BO) method against a widely-used non-BO baseline, 
Random Search. Random Search is a simple, uninformed optimization method that samples 
candidate solutions uniformly at random from the search space \cite{bergstra2012random}. 
Despite its simplicity, Random Search can perform competitively in some cases and 
serves as a strong baseline for assessing the effectiveness of more sophisticated 
optimization methods.

\section{Experimental Setup}

To evaluate the performance of QS-BO, we considered synthetic black-box optimization 
problems that are standard in the Bayesian optimization literature. Specifically, 
we optimized: 
\begin{itemize}
    \item a one-dimensional sinusoidal-quadratic function,
    \item the one-dimensional Forrester function, and
    \item the two-dimensional Branin function.
\end{itemize}
These functions are chosen for their non-trivial structure, which requires balancing 
exploration and exploitation during optimization, and are widely used benchmarks 
for global optimization algorithms.  

\subsection{1D Sinusoidal-Quadratic Function}
\leavevmode

The first test function is defined as
\[
f(x) = \sin(3x) + x^2 - 0.7x,
\]
with the search domain $x \in (-2.0, 2.0)$.  
This function is multimodal and non-convex, making it a suitable benchmark for 
demonstrating the trade-off between exploration and exploitation.  

\subsection{Forrester Function}
\leavevmode

The Forrester function is given by
\[
f(x) = (6x - 2)^2 \sin(12x - 4),
\]
with the domain $x \in [0, 1]$.  
It is widely used in the Bayesian optimization literature because of its 
complex shape, with multiple local minima and a global minimum, posing 
a challenge to surrogate-based optimization algorithms.  

\subsection{Branin Function}
\leavevmode

The Branin function is defined as
\[
f(x_1, x_2) = a \Big(x_2 - b x_1^2 + c x_1 - r \Big)^2 + s(1 - t)\cos(x_1) + s,
\]
where $a = 1$, $b = 5.1/(4\pi^2)$, $c = 5/\pi$, $r = 6$, $s = 10$, and 
$t = 1/(8\pi)$. The domain was restricted to $[-5, 10] \times [0, 15]$.  
This function has three global minima of equal value, making it a classic and 
challenging test case for global optimization algorithms.  

\subsection{Algorithms Compared}
\leavevmode

We compared:
\begin{itemize}
    \item \textbf{Random Search:} At each iteration, candidate points were drawn 
    uniformly at random from the domain, and the best function value observed was tracked.
    \item \textbf{QS-BO (proposed):} The optimization began with $n_\text{init} = 5$ 
    random evaluations, after which the surrogate was updated iteratively for 
    $n_\text{iter} = 30$ steps. At each iteration, 5,000 candidate points were sampled, 
    ranked, and transformed into quantiles before fitting a Gaussian process surrogate. 
    Standard acquisition functions were then applied to propose the next evaluation point.
\end{itemize}
Both methods were evaluated using the same total budget 
($n_\text{init} + n_\text{iter} = 35$).  

\subsection{Evaluation Protocol}
\leavevmode

We conducted $n_\text{runs} = 20$ independent runs with different random seeds 
for each method on each test function.  
For each run, the reported outcome is the lowest function value found after the 
allocated budget. Results across runs were summarized using descriptive statistics 
(mean, median, standard deviation, minimum, and maximum).  
To assess statistical significance, we applied both a two-sample $t$-test and 
the Wilcoxon signed-rank test.

\section{Results}
\begin{figure}[H]
    \centering
    \begin{subfigure}[t]{0.48\textwidth}
        \centering
        \includegraphics[width=\textwidth]{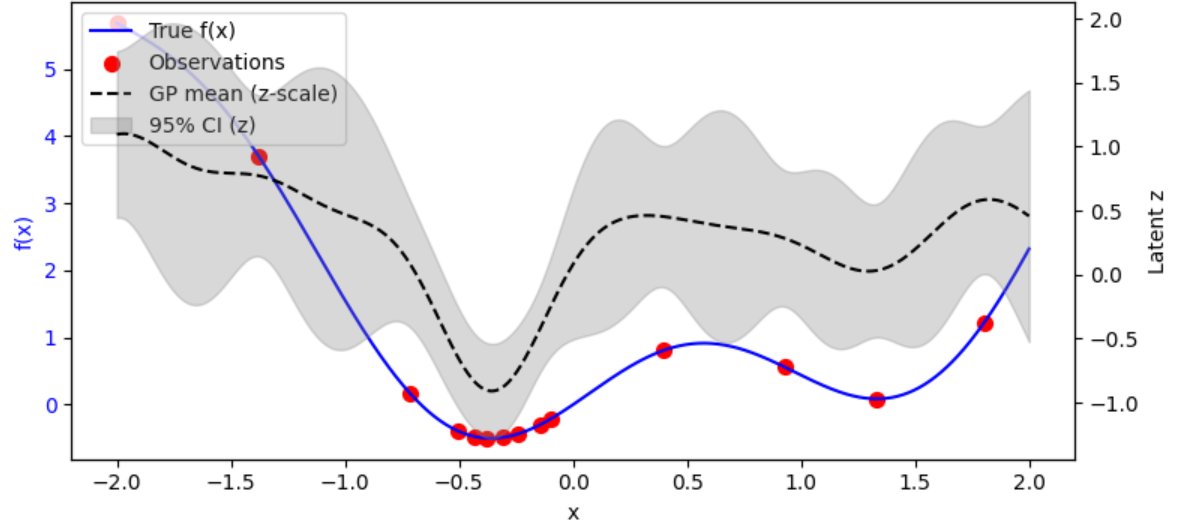}
        \caption{$f(x) = \sin(3x) + x^2 - 0.7x$}
        \label{fig:sub1}
    \end{subfigure}
    \hfill
    \begin{subfigure}[t]{0.48\textwidth}
        \centering
        \includegraphics[width=\textwidth]{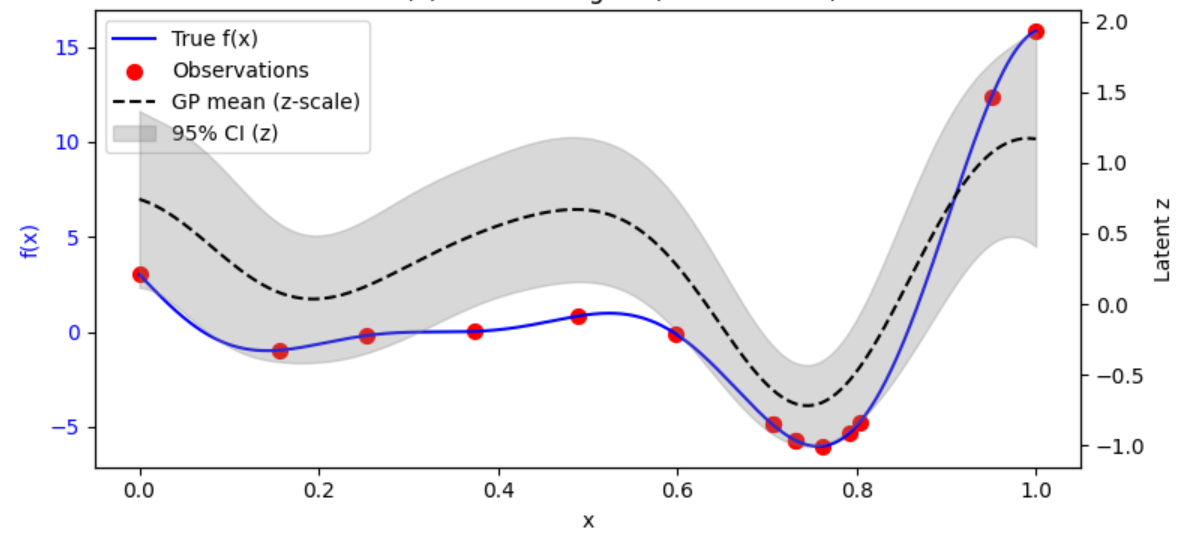}
        \caption{Forrester function: $(6x - 2)^2\sin(12x -4)$}
        \label{fig:sub2}
    \end{subfigure}
    
    \caption{Plots of 1D functions. 
    The latent Gaussian process model (dotted lines) closely follows the shape of the true function (blue lines), 
    and the evaluated points become increasingly concentrated around the minimum values of $f$.}
    \label{fig:1d_benchmarks}
\end{figure}

\begin{figure}[H]
    \centering
    \begin{subfigure}{0.32\textwidth}
        \centering
        \includegraphics[width=\linewidth]{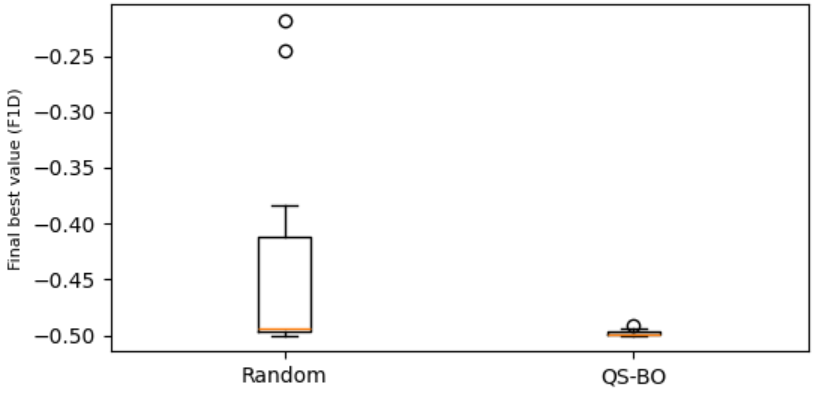}
        \caption{1D Sinusoidal–Quadratic}
        \label{fig:f1D_boxplot}
    \end{subfigure}
    \hfill
    \begin{subfigure}{0.32\textwidth}
        \centering
        \includegraphics[width=\linewidth]{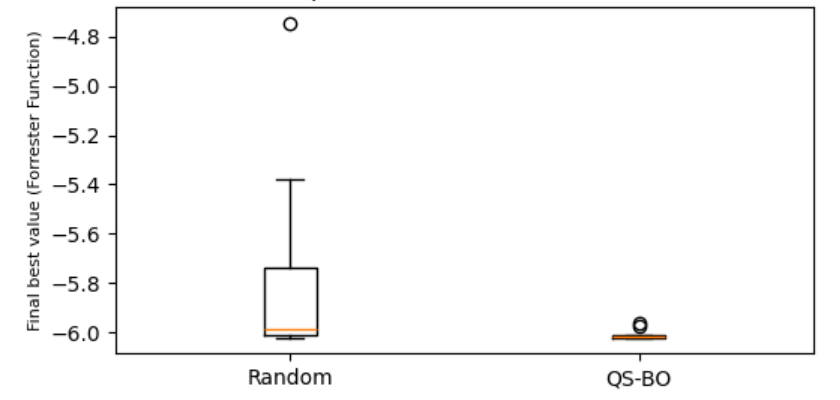}
        \caption{Forrester}
        \label{fig:forrester_boxplot}
    \end{subfigure}
    \hfill
    \begin{subfigure}{0.32\textwidth}
        \centering
        \includegraphics[width=\linewidth]{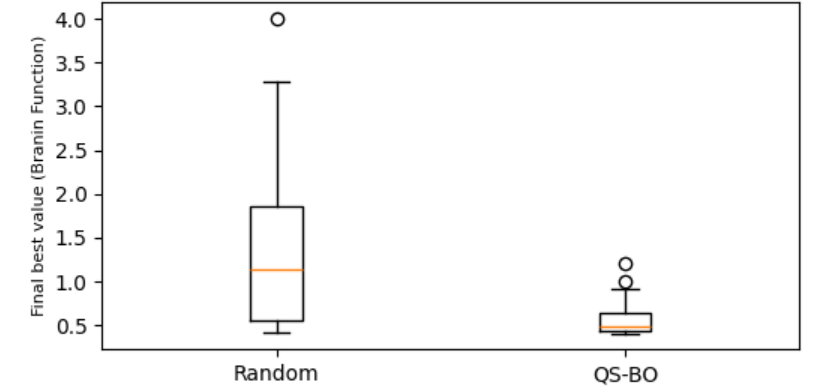}
        \caption{Branin}
        \label{fig:branin_boxplot}
    \end{subfigure}
    \caption{Boxplots comparing the performance of Random Search and QS-BO across multiple runs on three benchmark functions: (a) the 1D sinusoidal–quadratic function, (b) the Forrester function, and (c) the Branin function. In all cases, QS-BO achieves consistently lower function values than Random Search, highlighting its effectiveness.}
    \label{fig:all_boxplots}
\end{figure}

Across all runs, QS-BO consistently achieved lower final values compared to Random Search, 
with reduced variability. Both statistical tests confirmed that these improvements 
were significant at the 1\% level. Figure~\ref{fig:all_boxplots} shows a boxplot 
comparison of the two methods. QS-BO not only achieved lower objective values on average 
but also displayed a tighter distribution of results across runs, reflecting more stable 
performance. The visualization of the 1D functions \ref{fig:1d_benchmarks} shows the behaviour of the latent value against the actual function

\begin{table}[H]
    \centering
    \caption{Summary statistics of the performance of Random Search and QS-BO across benchmark functions. Reported values are aggregated over 20 independent runs.}
    \label{tab:summary_stats}
    \begin{tabular}{llccccc}
        \hline
        Function & Method & Mean & Median & Std.\ Dev. & Min & Max \\
        \hline
        1D Sinusoidal-Quadratic 
        & Random & -0.4431 & -0.4934 & 0.0813 & -0.5001 & -0.2181 \\
        & QS-BO  & -0.4980 & -0.4991 & 0.0027 & -0.5003 & -0.4909 \\
        \hline
        Forrester 
        & Random & -5.8176 & -5.9857 & 0.3097 & -6.0191 & -4.7465 \\
        & QS-BO  & -6.0117 & -6.0180 & 0.0153 & -6.0207 & -5.9635 \\
        \hline
        Branin 
        & Random & 1.4495 & 1.1444 & 1.0104 & 0.4190 & 4.0076 \\
        & QS-BO  & 0.5846 & 0.4777 & 0.2233 & 0.4030 & 1.2021 \\
        \hline
    \end{tabular}
\end{table}

\begin{table}[H]
    \centering
    \caption{Statistical significance tests comparing QS-BO against Random Search across benchmark functions. Both the paired $t$-test and Wilcoxon signed-rank test indicate that QS-BO significantly outperforms Random Search.}
    \label{tab:significance_tests}
    \begin{tabular}{lcccc}
        \hline
        Function & $t$-stat & $p$-value (t-test) & $W$ & $p$-value (Wilcoxon) \\
        \hline
        1D Sinusoidal-Quadratic & -2.94 & 0.0055 & 25.0 & 0.0017 \\
        Forrester               & -2.73 & 0.0096 & 15.0 & 0.0003 \\
        Branin                  & -3.64 & 0.0008 & 16.0 & 0.0003 \\
        \hline
    \end{tabular}
\end{table}


\section{Discussion}

In this work, we introduced Quantile-Scaled Bayesian Optimization (QS-BO), 
a rank-based optimization framework that extends the flexibility of Bayesian 
Optimization to settings where only rank feedback is available. By converting 
ranks into heteroscedastic Gaussian targets through a principled quantile-scaling 
pipeline, QS-BO allows the use of standard Gaussian process surrogates and 
acquisition functions without requiring access to explicit objective values. 
This provides a general and practical approach to optimization under preference 
or ordinal feedback.

Our experiments on one-dimensional nonlinear functions and the two-dimensional Branin function showed that QS-BO consistently outperformed Random Search, achieving lower final values and more stable performance. Boxplots (Fig \ref{fig:all_boxplots}) and summary statistics (Table \ref{tab:summary_stats}) confirmed these improvements, while statistical tests (Table \ref{tab:significance_tests}) established significance at the 1\% level. Visualizations further revealed that QS-BO adaptively concentrates evaluations near regions of low function values, demonstrating that preference information alone can guide effective global optimization.

Despite its promise, QS-BO faces challenges. Its reliance on Gaussian processes limits scalability to higher dimensions, and the rank-to-quantile transformation introduces additional computational cost compared to conventional BO. Addressing these challenges is essential for broader adoption.

Overall, our findings establish QS-BO as a principled and effective extension 
of Bayesian Optimization to rank-based settings, bridging the gap between 
preference learning and global optimization, and paving the way for broader 
applications where absolute objective values are unavailable or unreliable.

%% file: chapter5.tex
\chapter{Conclusion and Future Work}

\section{Conclusion}

This study demonstrated that QS-BO can effectively optimize black-box functions using only rank feedback. Across benchmark problems, QS-BO reliably identified better solutions than Random Search, highlighting the viability of ordinal information as a powerful signal for global optimization.


\section{Recommendations for Future Work}

Several directions remain open for further exploration:  

\begin{itemize}

    \item \textbf{Application to Hyperparameter Optimization:} While hyperparameter optimization motivated this work, future studies should extend QS-BO to real ML tasks (e.g., tuning neural networks, random forests, or autoregressive models) to validate performance in higher-dimensional, noisy settings.  
    \item \textbf{Comparison with Other Preference-Based Methods:} Benchmarking QS-BO against other ordinal- and preference-based BO variants would provide a clearer understanding of its relative strengths and weaknesses.  
    \item \textbf{Scalability to High Dimensions:} Current experiments focused on simple one-and-two-dimensional functions. Extensions to higher-dimensional search spaces, possibly with dimensionality reduction or additive kernel structures, should be explored.  
    \item \textbf{Incorporating Human-in-the-Loop Feedback:} Since ordinal judgments often arise from human evaluations, integrating QS-BO in settings such as recommender systems, A/B testing, or subjective quality assessments could broaden its applicability.  
    \item \textbf{Theoretical Analysis:} Formal guarantees on convergence rates and regret bounds under rank-only feedback remain an open theoretical challenge and merit further investigation.  
\end{itemize}